\documentclass{article}

\usepackage{microtype}
\usepackage{graphicx}
\usepackage{subcaption}
\usepackage{booktabs} %

\usepackage{hyperref}

\usepackage[preprint]{icml2026}

\usepackage{amsmath}
\usepackage{amssymb}
\usepackage{mathtools}
\usepackage{amsthm}
\usepackage{tcolorbox}

\usepackage{afterpage}

\usepackage[capitalize,noabbrev]{cleveref}

\theoremstyle{plain}

\theoremstyle{definition}

\theoremstyle{remark}

\usepackage[textsize=tiny]{todonotes}

\icmltitlerunning{JEPA-VLA: Video Predictive Embedding is Needed for VLA Models}

\begin{document}

\twocolumn[
  \icmltitle{JEPA-VLA: Video Predictive Embedding is Needed for VLA Models}

  \icmlsetsymbol{equal}{*}

  \begin{icmlauthorlist}
    \icmlauthor{Shangchen Miao}{tsinghua}
    \icmlauthor{Ningya Feng}{tsinghua}
    \icmlauthor{Jialong Wu}{tsinghua}
    \icmlauthor{Ye Lin}{tsinghua}
    \icmlauthor{Xu He}{huawei}
    \icmlauthor{Dong Li}{huawei}
    \icmlauthor{Mingsheng Long}{tsinghua}
  \end{icmlauthorlist}

  \icmlaffiliation{tsinghua}{Tsinghua University}
  \icmlaffiliation{huawei}{Huawei Noah's Ark Lab}

  \icmlcorrespondingauthor{Mingsheng Long}{mingsheng@tsinghua.edu.cn}

  \icmlkeywords{Machine Learning, ICML}

  \vskip 0.3in
]

\printAffiliationsAndNotice{}  %

\begin{abstract}

Recent vision-language-action (VLA) models built upon pretrained vision-language models (VLMs) have achieved significant improvements in robotic manipulation. However, current VLAs still suffer from low sample efficiency and limited generalization. This paper argues that these limitations are closely tied to an overlooked component, pretrained visual representation, which offers insufficient knowledge on both aspects of \textit{environment understanding} and \textit{policy prior}. Through an in-depth analysis, we find that commonly used visual representations in VLAs, whether pretrained via language-image contrastive learning or image-based self-supervised learning, remain inadequate at capturing crucial, task-relevant environment information and at inducing effective policy priors, i.e., anticipatory knowledge of how the environment evolves under successful task execution. In contrast, we discover that predictive embeddings pretrained on videos, in particular V-JEPA~2, are adept at flexibly discarding unpredictable environment factors and encoding task-relevant temporal dynamics, thereby effectively compensating for key shortcomings of existing visual representations in VLAs. Building on these observations, we introduce JEPA-VLA, a simple yet effective approach that adaptively integrates predictive embeddings into existing VLAs. Our experiments demonstrate that JEPA-VLA yields substantial performance gains across a range of benchmarks, including LIBERO, LIBERO-plus, RoboTwin2.0, and real-robot tasks.

\end{abstract}

\begin{figure*}[ht]
  \vskip 0.2in
  \begin{center}
    \centerline{\includegraphics[width=\columnwidth * 2]{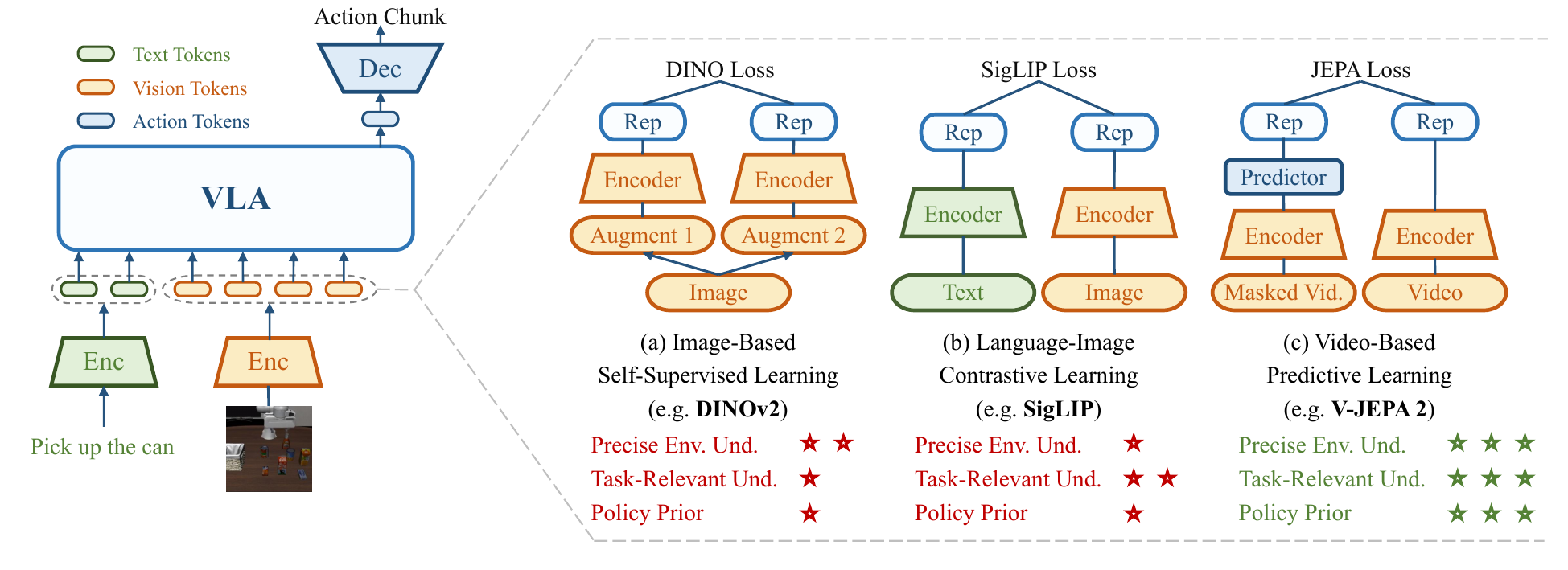}}
    \caption{
\textbf{Comparison of visual representations commonly used in VLAs:}
\textbf{(a) Image-based self-supervised learning (e.g., the DINO family)} yields precise visual representations, but is relatively insensitive to task relevance and preserves task-irrelevant details. 
\textbf{(b) Language-image contrastive learning (e.g., CLIP and the SigLIP family)} emphasizes instruction-aligned entities and semantics, yet may capture less low-level task-relevant information beyond what is explicitly described in text. 
\textbf{(c) Video-based predictive learning (e.g., V-JEPA~2)} provides state-centric representations for task-relevant objects while also encoding temporal regularities that act as policy priors, which are difficult to obtain from image-only pretraining.
    }
    \label{fig:main}
  \end{center}
  \vspace{-10pt}
\end{figure*}

\section{Introduction}\label{sec:intro}

Recent advances in Vision-Language-Action (VLA) models have significantly improved the integration of visual perception, natural language understanding, and action generation, enabling general-purpose agents to interpret and execute human instructions across diverse environments~\cite{kim2025fine, cen2025rynnvla}. These VLAs typically build upon large-scale pretrained Vision-Language Models (VLMs) by incorporating action heads or other specialized action generation modules. Despite these advances, current VLAs still struggle with low sample efficiency and limited generalization to new scenarios. Many VLAs are trained on millions of trajectories~\cite{bjorck2025gr00t, zitkovich2023rt, qu2025spatialvla}, yet cover only tens of tasks, and they can experience substantial performance degradation--sometimes up to $\sim$40\%--when evaluated on unseen tasks or under distribution shifts~\cite{sapkota2025vision}.

We argue that these limitations are closely tied to bottlenecks in visual understanding imposed by VLM's vision backbones or other visual processors~\cite{sapkota2025vision, shao2025large}. In particular, robotics requires two critical forms of visual knowledge: (i) \textit{environment understanding}, which precisely captures task-relevant object attributes (e.g., coordinates of targets) while flexibly discarding task-irrelevant information (e.g., lighting conditions); (ii) \textit{policy priors}, which encode anticipatory knowledge of how the environment evolves during successful task execution, thereby guiding action learning toward favorable future states.

However, visual representations used in most VLMs and VLAs are pretrained on large-scale datasets via either image-based self-supervised learning (e.g., the DINO family~\cite{caron2021emerging, oquab2024dinov2, simeoni2025dinov3}) or language-image contrastive learning (e.g., CLIP \cite{radford2021learning} and the SigLIP family~\cite{zhai2023sigmoid, tschannen2025siglip}). These representations exhibit notable deficiencies in providing the visual knowledge required for robotic control. First, they can be suboptimal for \emph{environment understanding}. Image-level self-supervised objectives like DINO encourage invariance to a broad set of augmentations, which may introduce inductive biases that are misaligned with manipulation tasks. For instance, invariance to random crops can reduce sensitivity to object positions and spatial configurations.  In contrast, language-image contrastive objectives emphasize semantics grounded in text-referred entities, but are often less effective at preserving other task-relevant cues (e.g., obstacles not mentioned in instructions, captions, or descriptions). Second, these representations provide weak \emph{policy priors}. Since they are typically learned from single images, they remain largely static and fail to adequately capture the temporal dynamics of successful action execution reflected in pretraining data, which are crucial for effective policy learning.

Motivated by these insights, we turn to predictive embeddings \cite{lecun2022path} learned from videos. In particular, we consider V-JEPA~2~\cite{assran2025v}, a joint-embedding predictive architecture pretrained on internet-scale videos. By predicting masked video patches in a latent space, V-JEPA~2 encourages representations that emphasize predictable, task-relevant factors while suppressing unpredictable nuisances. As a result, it is better suited to encoding dynamic cues about objects and agents (e.g., motion-relevant signals) that are critical for understanding the environment in robotics. Moreover, video-based pretraining allows V-JEPA~2 to internalize temporal regularities in how scenes evolve over successful task executions, which can serve as effective policy priors---a capability that is difficult to obtain from static, image-only pretraining.

To substantiate these insights, we conduct in-depth analyses on underlying state estimation and prediction. Notably, V-JEPA~2 outperforms all prior representations in these tasks (see Section~\ref{sec:analysis_of_features}), indicating its superior ability to capture precisely task-relevant environment information and crucial temporal regularities for robotic decision making. Building on these findings, we propose \textit{JEPA-VLA}, a simple yet effective approach to incorporate V-JEPA~2 representations into VLAs, thereby strengthening their knowledge of both environment understanding and policy priors. We evaluate our approach across multiple benchmarks, including LIBERO~\cite{liu2023libero}, LIBERO-plus~\cite{fei2025libero}, RoboTwin2.0~\cite{chen2025robotwin}, and a real-world experiment, demonstrating consistent and substantial improvements in task performance.

The main contributions of this work are threefold:
\begin{itemize}
    \item We identify two essential aspects of visual knowledge required by VLAs---environment understanding and policy priors---and show that commonly used static visual representations do not adequately provide either.
    \item We demonstrate that V-JEPA~2, a video-based predictive representation, effectively captures both knowledge and outperforms image-based and language-image-based representations in our analysis tasks.
    \item We propose JEPA-VLA, a simple and general framework for integrating video-predictive visual representations into existing VLAs, yielding consistent improvements in sample efficiency and generalization across multiple benchmarks and real-world tasks.
\end{itemize}

\begin{figure*}[ht]
  \vskip 0.2in
  \begin{center}
    \centerline{\includegraphics[width=\textwidth]{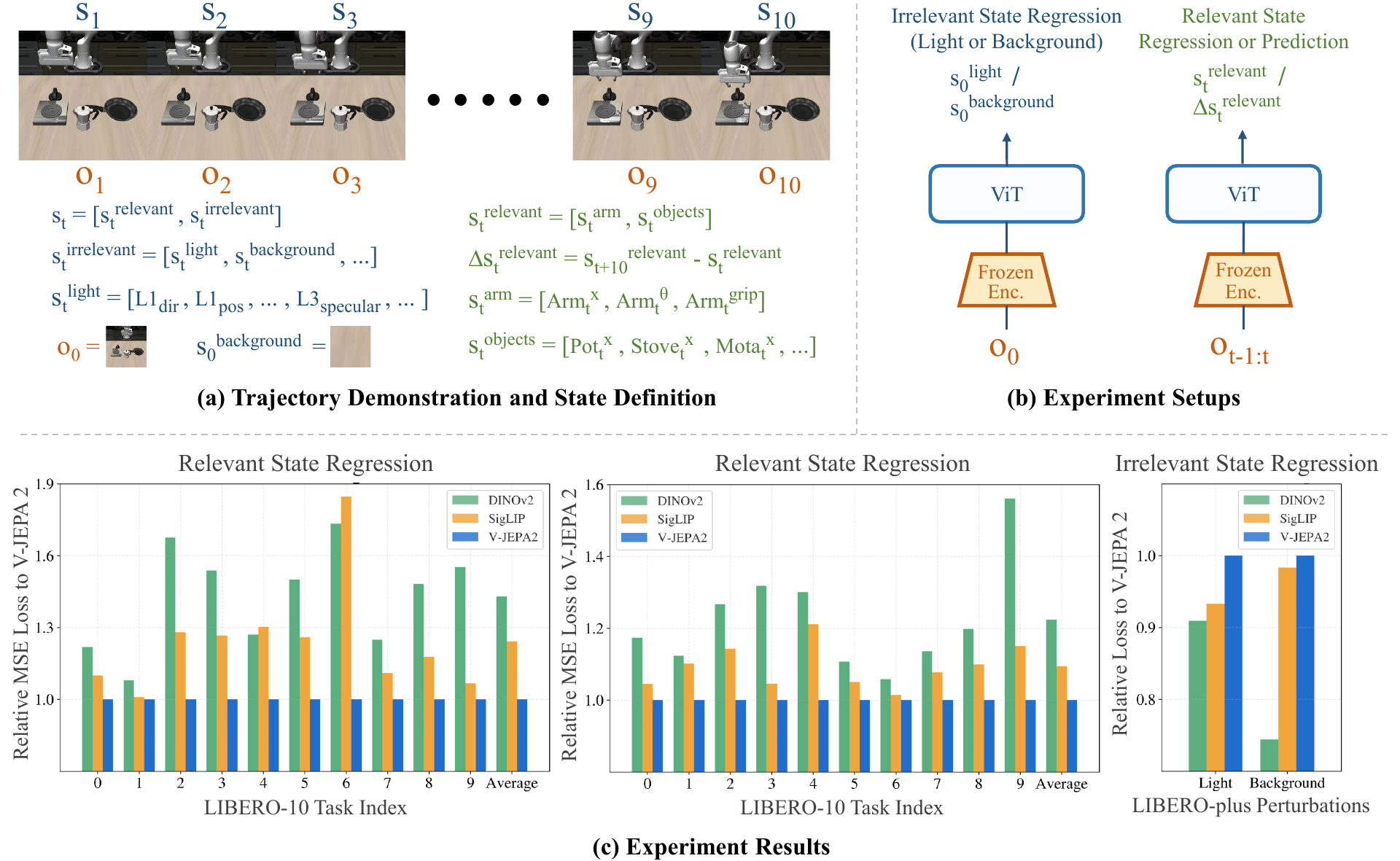}}
    \caption{\textbf{Analysis of visual representations for VLAs.}
    \textbf{(a) Trajectory demonstration and state definition.} Given observations $\{o_t\}$, we factorize the underlying state at each timestep as task-relevant and task-irrelevant parts.
    \textbf{(b) Experimental setup.} We freeze different vision encoders and train a lightweight ViT head to regress or predict environment states. More details can be found in Appendix~\ref{appendix:analysis}.
    \textbf{(c) Experimental results.} V-JEPA~2 achieves consistently lower relative loss on task-relevant regression and prediction, compared to DINOv2 and SigLIP, while showing no advantage on task-irrelevant (lighting/background) regression, suggesting that V-JEPA~2 better captures task-relevant environment states and policy priors while discarding nuisance factors.}
    \label{fig:analyse_architecture}
  \end{center}
\end{figure*}

\section{An Analysis of Vision Representations for VLAs}\label{sec:analysis_of_features}

We denote trajectories of robotic manipulation tasks as a partially observable Markov decision process (POMDP), formulated as a tuple $\langle \mathcal{S}, \mathcal{O}, \phi, \mathcal{A}, p \rangle$. At each time step, $s_t \in \mathcal{S}$ denotes the underlying environment state, while the agent only observes $o_t = \phi(s_t)$, which provides incomplete information about $s_t$. Given an action $a_t \sim \pi(o_{1:t})$, the environment transitions to a new state $s_{t+1} \sim p(s_{t+1} \mid s_t, a_t)$. This formulation highlights that effective robotic manipulation critically depends on inferring latent environment states and their evolution from partial visual observations.

Effective VLAs must therefore embrace two forms of knowledge from visual representations. First, they must infer task-relevant environment states from partial observations, which can be formalized as learning an internal state estimator $q_\theta(s_t \mid o_{1:t})$. Second, they must acquire policy-aligned anticipatory knowledge that captures how states evolve under successful action $a^{*}$ when completing specific tasks, i.e., $p_\theta(s_{t+1} \mid s_t, a_{t}^{*})$, which in turn induces policy priors. This perspective is related to recent work on video generation models as policies \cite{du2023learning, feng2025vidar}, which leverage large-scale video generative pretraining to transfer temporal regularities as inductive biases for control. By focusing on visual representations, our approach requires only minimal, plug-and-play modifications to existing VLAs and may benefit from the efficiency of focusing on predictable aspects while ignoring unpredictable details that generative objectives emphasize.

As discussed earlier, most visual encoders in current VLAs are pretrained on static images and provide insufficient support for policy learning with either form of knowledge. In contrast, we adopt V-JEPA~2 and hypothesize that its video-based predictive objective enables the encoding of temporal regularities that simultaneously support accurate environment understanding and induce effective policy priors.

In the remainder of this section, we conduct a comparative analysis to examine whether different visual representations (i) understand the current environment state, capturing crucial task-relevant information while discarding nuisance factors, and (ii) offer effective policy priors that anticipate how the environment evolves under successful actions. An overview of these experiments is shown in Figure ~\ref{fig:analyse_architecture}.

\subsection{Environment Understanding}

Effective manipulation requires visual representations to capture task-relevant aspects of the environment, such as the states of the robot arm and movable objects, while discarding nuisance factors that do not affect action execution (e.g., background texture and lighting). To evaluate whether different visual representations exhibit these properties, we design two probing experiments that separately assess their ability to encode task-relevant states and to ignore task-irrelevant variations.

\subsubsection{Task-Relevant State Estimation}
\label{sec:env_state}

\paragraph{Setup} 
We conduct experiments on the LIBERO-10 benchmark \cite{liu2023libero} by regressing the states of the robot arm and task-relevant objects from visual representations. The vision encoders are frozen, and a lightweight regression head is trained using representations extracted from the most recent two frames. We use the mean squared error (MSE) as the evaluation metric.

\vspace{-5pt}
\paragraph{Results} 
As shown in Figure~\ref{fig:analyse_architecture}c \textit{(left)}, V-JEPA~2 achieves a lower MSE when regressing current task-relevant states compared to DINOv2 and SigLIP. 
Image-based self-supervised representations such as DINOv2 tend to preserve detailed visual information across the entire image, which may introduce inductive biases misaligned with manipulation tasks, resulting in generally precise but not task-focused representations.
In contrast, language--image representations like SigLIP emphasize entities referred to in the text but may overlook task-relevant objects that are not commonly explicitly mentioned.
V-JEPA~2, pretrained with video-based predictive objectives, captures object-centric and motion-related cues, providing representations more aligned with task-relevant states.

\begin{tcolorbox}[colback=blue!2!white,leftrule=2.5mm,size=title]
\textbf{Finding 1.} \textit{V-JEPA~2 representations encode task-relevant environment states more precisely than DINOv2 and SigLIP.}
\end{tcolorbox}

\subsubsection{Task-Irrelevant State Estimation}

\paragraph{Setup} 
We further evaluate the sensitivity of visual representations to task-irrelevant factors using the LIBERO-plus benchmark \cite{fei2025libero}, which introduces perturbations that do not affect the required action sequence. We focus on lighting and background variations, and train models to regress these perturbation attributes from visual representations. Since such perturbations remain constant within each trajectory, only the first frame is used as input.

\vspace{-5pt}
\paragraph{Results} 
As shown in Figure~\ref{fig:analyse_architecture}c \textit{(right)}, V-JEPA~2 exhibits higher error in regressing lighting and background perturbations compared to DINOv2 and SigLIP. This result indicates that V-JEPA~2 encodes less information about task-irrelevant nuisance factors, suggesting that its representations selectively focus on dynamics and visual cues that are more relevant to manipulation. Such insensitivity to irrelevant variations is desirable and contributes to improved robustness and generalization.

\begin{tcolorbox}[colback=blue!2!white,leftrule=2.5mm,size=title]
\textbf{Finding 2.} \textit{V-JEPA~2 representations encode less information about task-irrelevant visual factors, compared to DINOv2 and SigLIP.}
\end{tcolorbox}

\subsection{Policy Priors}

Policy priors correspond to anticipatory knowledge of how task-relevant environment states are expected to evolve under successful action execution. Rather than directly predicting future observations, we evaluate whether visual representations encode crucial transition-aware information that reflects changes in task-relevant states despite partial observability.

\vspace{-5pt}
\paragraph{Setup} 
We train models on the LIBERO-10 benchmark to predict future state changes. Specifically, to mitigate the effects of partial observations on inaccurate state estimation, we train a lightweight head to predict the residual between the task-relevant state 10 steps later and the current state, using frozen visual representations.

\vspace{-5pt}
\paragraph{Results} 
As shown in Figure~\ref{fig:analyse_architecture}c (middle), V-JEPA~2 achieves lower MSE in predicting state residuals compared to DINOv2 and SigLIP. This suggests that V-JEPA~2 encodes transition-aware representations that capture temporal regularities aligned with successful task execution. In contrast, image-based representations pretrained on static images lack explicit temporal supervision and struggle to capture such transition dynamics, limiting their ability to induce effective policy priors.

\begin{tcolorbox}[colback=blue!2!white,leftrule=2.5mm,size=title]
\textbf{Finding 3.} \textit{V-JEPA~2 representations effectively embed policy priors.}
\end{tcolorbox}
\section{JEPA-VLA}\label{sec:method}

In light of these findings, we propose JEPA-VLA, a simple yet effective approach for incorporating powerful V-JEPA~2 representations into VLAs.

\subsection{Problem Formulation}

We consider two main components: an action model \(\pi_{\theta}\) and a frozen, pretrained V-JEPA~2 encoder \(E_{\phi}\). The action model \(\pi_{\theta}\) follows a standard VLA formulation: it takes as input a language instruction \(l\), multi-view observations from \(N\) cameras (e.g., head, wrist, and auxiliary views) \(o_{t}^{0:N}\), and the robot’s proprioceptive state \(s_{t}\), and produces an action \(a_{t}\) according to
\begin{equation}
a_{t} \sim \pi_{\theta}(a_{t} \mid l, o_{1:t}^{0:N}, s_{t}).
\end{equation}

Meanwhile, the frozen V-JEPA~2 encoder is built on a ViT~\cite{dosovitskiy2020image} backbone and maps a video clip \(o_{t-h:t} \in \mathbb{R}^{T \times H \times W \times 3}\) to visual representations \(h_{t} \in \mathbb{R}^{N \times C}\), where \(N\) scales with the spatial and temporal resolution \(H\), \(W\), and \(T\). This process is formalized as
\begin{equation}
h_{t} \sim E_{\phi}(h_{t} \mid o_{t-h:t}).
\end{equation}

We aim to learn an enhanced action model that effectively leverages the predictive visual representation \(h_{t}\):
\begin{equation}
a_{t} \sim \pi_{\theta}(a_{t} \mid l, o_{1:t}^{0:N}, s_{t}, h_{t}),
\end{equation}
thereby leading to improved environment understanding and more reliable action generation.

\subsection{Representation Fusion}

To condition the action model on additional visual representations, we instantiate two fusion strategies.

\paragraph{Early Fusion} Most VLAs are built on Transformer backbones. A natural approach is to treat V-JEPA~2 representations as \emph{additional input embeddings} and concatenate them with the original token sequence. This design is lightweight, as it introduces only a linear projection to align representation dimensions before fusion. Empirically, this simple concatenation works well for VLAs \emph{without} large-scale robot-manipulation pretraining, where the policy is largely learned from scratch (despite sharing common VLM priors).

\paragraph{Gated Fusion} However, this naive fusion is ineffective for VLAs pretrained on large-scale robot-manipulation data. Directly injecting extra tokens can shift the input distribution and interfere with the pretrained representations, leading to degraded transfer. To better adapt V-JEPA~2 representations to this more commonly used and data-efficient setting, we design an alternative architecture that preserves pretrained priors. Inspired by Flamingo~\cite{alayrac2022flamingo}, we incorporate V-JEPA~2 representations via several \emph{gated cross-attention} layers, where the original VLA tokens serve as queries, and the V-JEPA~2 representations serve as keys and values. This gated design allows the VLA to selectively attend to predictive embeddings when beneficial, enabling adaptive integration while minimally disrupting pretrained knowledge.

In practice, to balance learning performance, memory usage, and inference latency, we do not insert a gated cross-attention layer after every transformer decoder layer. Instead, guided by empirical results, we adopt a sparse fusion scheme that inserts gated cross-attention at a fixed interval across the decoder stack, which we find to be both efficient and sufficiently effective. To stably incorporate complementary information from V-JEPA~2 without disrupting pretrained VLA representations, we follow the design principles of Flamingo~\cite{alayrac2022flamingo} to set the learning rate of the newly introduced fusion layers substantially lower than that of the original VLA parameters. Further architectural details are provided in Appendix~\ref{appendix:jepa_vla}.

\begin{figure*}[!t]
  \vskip 0.2in
  \begin{center}
    \vspace{-10pt}
    \centerline{\includegraphics[width=\textwidth]{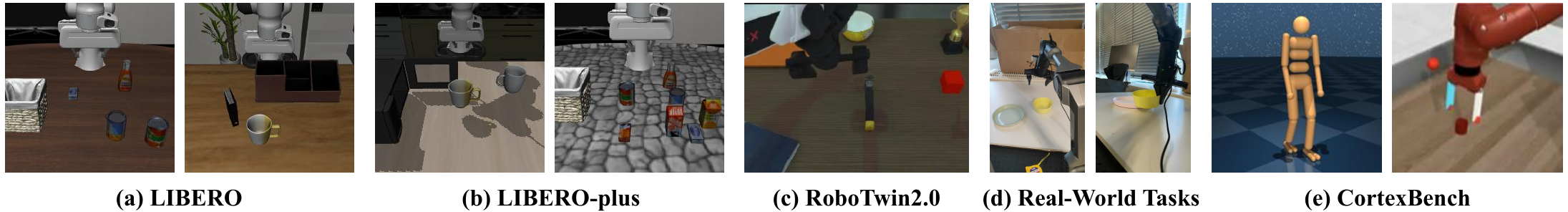}}
    \caption{
        Evaluation benchmarks, including (a) LIBERO, (b) LIBERO-plus, (c) RoboTwin, (d) a real-world task, and (e) CortexBench.
    }
    \vspace{-10pt}
    \label{fig:benchmark_showcase}
  \end{center}
\end{figure*}

\begin{table*}[!t]
    \caption{\textbf{LIBERO task success rate (\%) in the basic VLA setting.} 
    \emph{Baseline} denotes the basic VLA without V-JEPA~2 fusion.}
    
  \label{tab:libero1-10}
  \begin{center}
    \begin{small}
      \begin{sc}
        \begin{tabular}{lcccccr}
          \toprule
          Model~ & ~Spatial~ & ~Object~ & ~Goal~ & ~Long~ & ~Average \\
          \midrule
          Baseline    & 58.2 & 74.8 & 87.8 & 25.8 & 61.65 \\
          \textbf{w/ ours}  & \textbf{69.2} & \textbf{78.2} & \textbf{87.8} & \textbf{41.0} & \textbf{69.05}\\
          $\Delta$  & \textcolor[RGB]{17,159,73}{+11.0} & \textcolor[RGB]{17,159,73}{+3.4} & \textcolor[RGB]{100,100,100}{0.0} & \textcolor[RGB]{17,159,73}{+15.2} & \textcolor[RGB]{17,159,73}{+7.40}\\
          \bottomrule
        \end{tabular}
      \end{sc}
    \end{small}
  \end{center}
  \vskip -0.1in
\end{table*}

\begin{table*}[!t]
    \caption{\textbf{LIBERO-plus task success rate (\%) in the basic VLA setting.} 
    \emph{Baseline} denotes the Basic VLA without V-JEPA~2 fusion.}
  \label{tab:libero_plus}
  \begin{center}
    \begin{small}
      \begin{sc}
        \begin{tabular}{lccccccccr}
          \toprule
          Model & Camera & Robot & Language & Light & Background & Noise & Layout & Total \\
          \midrule
          WorldVLA & 0.1 & 27.9 & 41.6 & 43.7 & 17.1 & 10.9 & 38.0 & 25.0\\
          Baseline & 0.1 & 20.8 & 43.9 & 26.3 & 10.1 & 4.1 & 26.9 & 18.9 \\
          \textbf{w/ ours}  & \textbf{0.4} & \textbf{25.7} & \textbf{54.4} & \textbf{38.0} & \textbf{23.9} & \textbf{4.1} & \textbf{32.8} & \textbf{25.6}\\
          $\Delta$  & \textcolor[RGB]{17,159,73}{+0.3} & \textcolor[RGB]{17,159,73}{+4.9} & \textcolor[RGB]{17,159,73}{+10.5} & \textcolor[RGB]{17,159,73}{+11.7} & \textcolor[RGB]{17,159,73}{+13.8} & \textcolor[RGB]{100,100,100}{0.0} & \textcolor[RGB]{17,159,73}{+5.9} & \textcolor[RGB]{17,159,73}{+6.7}\\
          \bottomrule
        \end{tabular}
      \end{sc}
    \end{small}
  \end{center}
  \vskip -0.1in
\end{table*}

\section{Experiments}\label{sec:experiments}

In this section, we investigate whether simply integrating V-JEPA~2 representations can consistently improve manipulation policies not only in in-domain, data-rich regimes, but also under out-of-domain shifts and limited-data settings. We describe the simulation and real-world benchmarks in Section~\ref{exp:benchmark} and provide implementation details in Section~\ref{exp:implementation}. Section~\ref{exp:evaluation_results} presents the benchmark results. Finally, in Section~\ref{exp:analysis}, we examine whether V-JEPA~2 outperforms other pretrained visual representations across a broader range of embodied AI tasks.

\subsection{Benchmarks}\label{exp:benchmark}

We validate JEPA-VLA across diverse environments, as shown in Figure~\ref{fig:benchmark_showcase}.

\textbf{LIBERO}~\cite{liu2023libero} is a benchmark comprising four task suites designed to study lifelong learning and knowledge transfer in robotics. Each suite contains 10 tasks with 50 tele-operated demonstrations. We use all four suites: \emph{LIBERO-Spatial} (spatial reasoning), \emph{LIBERO-Object} (object understanding), \emph{LIBERO-Goal} (instruction following), and \emph{LIBERO-Long} (long-horizon tasks with diverse layouts).

\textbf{LIBERO-plus}~\cite{fei2025libero} extends LIBERO by introducing controlled perturbations along multiple dimensions, including camera viewpoint, initial state, language instructions, lighting, background, noise, and layout.

\textbf{RoboTwin}~\cite{mu2025robotwin, chen2025robotwin} provides simulation benchmarks for dual-arm manipulation. We use \textbf{RoboTwin2.0}~\cite{chen2025robotwin} in our experiments. RoboTwin2.0 includes 50 tasks across multiple robot platforms and comprises 731 object instances spanning 147 categories. It also incorporates comprehensive domain randomization (e.g., clutter, lighting, background, tabletop height, and language instructions), requiring policies to remain robust under transfer. We adopt the Aloha-AgileX dual-arm setup and evaluate on six tasks.

\textbf{Real-World Task}~We evaluate on a single Piper robot arm. We design a pick-and-place task and train both the baseline and our method. To assess performance under limited data, we additionally train our method on a one-fifth subset of the training trajectories (22 trajectories). We conduct tests under various settings to verify whether JEPA-VLA can improve generalization. More settings and implementation details can be found in Appendix~\ref{appendix:real_world}.

\begin{table*}[!t]
  \caption{\textbf{LIBERO task success rate (\%) in the mainstream VLA setting.} $^{\dagger}$Results are taken from official paper~\cite{kim2025fine}.}
  \label{tab:libero_full}
  \begin{center}
    \begin{small}
      \begin{sc}
        \begin{tabular}{lcccccr}
          \toprule
          Model~ & ~Spatial~ & ~Object~ & ~Goal~ & ~Long~ & ~Average \\
          \midrule
          OpenVLA-OFT$^{\dagger}$ & 96.2 & 98.3 & 96.2 & 90.7 & 95.35 \\
          OpenVLA-OFT & 84.4 & 91.8 & 95.2 & 89.8 & 90.30 \\
          \textbf{w/ ours}  & \textbf{97.2} & \textbf{98.0} & \textbf{95.6} & \textbf{94.8} & \textbf{96.40}\\
          $\Delta$  & \textcolor[RGB]{17,159,73}{+12.8} & \textcolor[RGB]{17,159,73}{+6.2} & \textcolor[RGB]{17,159,73}{+0.4} & \textcolor[RGB]{17,159,73}{+5.0} & \textcolor[RGB]{17,159,73}{+6.10}\\
          \bottomrule
        \end{tabular}
      \end{sc}
    \end{small}
  \end{center}
  \vskip -0.1in
\end{table*}

\setlength{\tabcolsep}{3pt} 
\begin{table*}[!t]
\centering
\caption{\textbf{RoboTwin2.0 task success rate (\%) in the mainstream VLA setting.} \emph{Baseline} denotes our implementation of OpenVLA-OFT.}
\label{tab:robotwin_style}
\small
\begin{tabular}{l cc cc cc cc cc cc cc cc}
\toprule
& \multicolumn{2}{c}{Beat Block Hammer} & \multicolumn{2}{c}{Click Alarmclock} & \multicolumn{2}{c}{Lift Pot} & \multicolumn{2}{c}{Place Fan} & \multicolumn{2}{c}{Place Phone Stand} & \multicolumn{2}{c}{Stack Bowls Two} & \multicolumn{2}{c}{Average} \\
\cmidrule(lr){2-3}\cmidrule(lr){4-5}\cmidrule(lr){6-7}\cmidrule(lr){8-9}\cmidrule(lr){10-11}\cmidrule(lr){12-13}\cmidrule(lr){14-15}
Model & Easy & Hard & Easy & Hard & Easy & Hard & Easy & Hard & Easy & Hard & Easy & Hard & Easy & Hard \\
\midrule
Baseline & 18.0 & 0.0 & 85.0 & 26.0 & 93.0 & 18.0 & 25.0 & 0.0 & 33.0 & 4.0 & 75.0 & 8.0 & 54.8 & 9.3 \\
\textbf{Ours} & \textbf{65.0} & \textbf{13.0} & \textbf{92.0} & \textbf{30.0} & \textbf{98.0} & \textbf{21.0} & \textbf{38.0} & \textbf{0.0} & \textbf{60.0} & \textbf{16.0} & \textbf{88.0} & \textbf{26.0} & \textbf{73.5} & \textbf{17.7} \\
$\Delta$ & \textcolor[RGB]{17,159,73}{+47.0} & \textcolor[RGB]{17,159,73}{+13.0} & \textcolor[RGB]{17,159,73}{+7.0} & \textcolor[RGB]{17,159,73}{+4.0} & \textcolor[RGB]{17,159,73}{+5.0} & \textcolor[RGB]{17,159,73}{+3.0} & \textcolor[RGB]{17,159,73}{+13.0} & \textcolor[RGB]{100,100,100}{0.0} & \textcolor[RGB]{17,159,73}{+27.0} & \textcolor[RGB]{17,159,73}{+12.0} & \textcolor[RGB]{17,159,73}{+13.0} & \textcolor[RGB]{17,159,73}{+18.0} & \textcolor[RGB]{17,159,73}{+18.7} & \textcolor[RGB]{17,159,73}{+8.4} \\
\bottomrule
\end{tabular}
\end{table*}

\subsection{Implementation Details}\label{exp:implementation}

We implement JEPA-VLA in a standard VLA setting where the policy conditions on a third-person visual observation and a language instruction. Concretely, we extract V-JEPA~2 representations from the two most recent frames and use them as additional conditioning signals alongside the current observation when predicting actions. Two base VLAs are experimented with, as described below.

\paragraph{Basic VLA Experiments}\label{exp:worldvla_imple}  
To assess the effectiveness of V-JEPA~2 representations for VLAs, we begin by implementing a basic VLA model architecture, consisting of a pretrained Vision-Language Model (VLM) backbone and a linear action prediction head. Inspired by the architecture of WorldVLA~\cite{cen2025worldvla}, we use Chameleon~\cite{team2024chameleon} as the VLM backbone and discretize the actions into 256 tokens, as done in previous works~\cite{kim2025openvla}. We evaluate this setting on both LIBERO and LIBERO-plus. For both datasets, as this is an experimental trial, we train using only 1/10 of the LIBERO data. Additionally, we conduct trials with the same architecture for real-world experiments.

\paragraph{Mainstream VLA Experiments} 
We further evaluate our method by incorporating V-JEPA~2 representations into mainstream VLAs on LIBERO and RoboTwin2.0. We implement one of the standard OpenVLA-OFT~\cite{kim2025fine} configurations, using third-person images and language instructions as input, with parallel decoding, action chunking, and continuous action prediction, employing an $\ell_1$ regression loss. For LIBERO, we follow the official data preprocessing procedure provided by the OpenVLA-OFT implementation. For RoboTwin2.0, we train with the official 50 clean trajectories and evaluate under both clean and domain-randomized settings.We insert gated cross-attention layers every \emph{eight} decoder layers for both benchmarks.

\paragraph{Training Details}  

We follow the official baseline training settings whenever possible, adjusting only the batch size to accommodate limited computational resources. For LIBERO and LIBERO-plus, we train both our method and the baselines with a batch size of 16. The Basic VLA is trained for 40 epochs to ensure convergence, while OpenVLA-OFT is trained for 150k steps following the official schedule. For RoboTwin2.0, we use a batch size of 8 and train for approximately 100k steps. For the newly introduced fusion layers, we set a smaller learning rate of \(1 \times 10^{-5}\) to \(1 \times 10^{-4}\), compared to \(5 \times 10^{-4}\) for the original VLA parameters in OpenVLA-OFT.

\subsection{Main Results}\label{exp:evaluation_results}

As shown in Tables~\ref{tab:libero1-10} and~\ref{tab:libero_plus}, incorporating V-JEPA~2 representations into the basic VLA improves success rates by 7.4\% on LIBERO and 6.7\% on LIBERO-plus. Notably, on LIBERO-plus, our method outperforms the official WorldVLA model despite being trained with only one-tenth of the action-model data and without any world-model training data, indicating that V-JEPA~2 representations substantially enhance generalization (see Appendix~\ref{appendix:libero_plus} for per-task details). In real-world experiments (Table~\ref{tab:real_world}), V-JEPA~2 representations enable training with only one-fifth of the trajectories while still outperforming the baseline, further validating improved performance and generalization in deployment settings (trajectory examples are provided in Appendix~\ref{appendix:real_world}).

Our method also transfers effectively to mainstream VLA architectures. As shown in Table~\ref{tab:libero_full}, despite using a smaller batch size, our method improves over the baseline by 6.1\% and even surpasses the official OpenVLA-OFT results. On RoboTwin2.0 (Table~\ref{tab:robotwin_style}), our method achieves consistent average gains of 18.7\% and 8.4\% in the clean and domain-randomized settings, respectively. Overall, these results indicate that JEPA-VLA is not only effective in controlled, simplified evaluations but also yields robust improvements on mainstream VLA backbones, demonstrating the generality of our method.

\begin{table}[t]
  \caption{\textbf{Real-world task (pick the yellow bowl into the plate) success rate (\%)}. \emph{Layout} means different desktop item layouts, and \emph{light} means different brightness.}
  \label{tab:real_world}
  \begin{center}
    \begin{small}
      \begin{sc}
        \begin{tabular}{lcccr}
          \toprule
          Model & Average & Layout & Light  \\
          \midrule
          Baseline & 50.0 & 0.0 & 0.0 \\
          w / ours 1/5 data & 60.0 & 33.3 & 66.7\\
          \textbf{w/ ours} full data  & \textbf{80.0} & \textbf{100.0} & \textbf{100.0}\\
          \bottomrule
        \end{tabular}
      \end{sc}
    \end{small}
  \end{center}
  \vskip -0.1in
\end{table}

\begin{table*}[t]
\centering
\caption{\textbf{CortexBench results.} \emph{Top:} MetaWorld task success rate (\%). \emph{Bottom:} DMControl episode reward.}
\vspace{-5pt}

\label{tab:cortexbench_1}
\begin{center}
    \begin{small}
      \begin{sc}
        \begin{tabular}{lcccccr}
          \toprule
          MetaWorld & assembly & bin picking & button press & drawer open & hammer & average\\
          \midrule
          CLIP (VIT-B) & 70.7 & 68.0 & 48.0 & 100.0 & 90.7 & 75.5 \\
          VC-1 (VIT-L) & 88.0 & \textbf{84.0} & 80.0 & 100.0 & 92.0 & 88.8  \\
          \textbf{V-JEPA~2 (VIT-L)} & \textbf{96.0} & 80.0 & \textbf{84.0} & \textbf{100.0} & \textbf{92.0} & \textbf{90.4} \\
          \bottomrule
        \end{tabular}
       \end{sc}
   \end{small}
\end{center}
\begin{center}
    \begin{small}
        \begin{sc}
        \begin{tabular}{lcccccr}
          \toprule
          DMControl & walker stand & walker walk & reacher easy & cheetah run & finger spin & average\\
          \midrule
          VC-1 (VIT-L) & 811 & 338 & \textbf{836} & 155 & 540 & 536.0  \\
          \textbf{V-JEPA~2 (VIT-L)} & \textbf{825} & \textbf{436} & 821 & \textbf{156} & \textbf{681} & \textbf{583.8} \\
          \bottomrule
        \end{tabular}
      \end{sc}
    \end{small}
  \end{center}
  \vspace{-5pt}
\end{table*}

\subsection{Comparative Evaluation of Visual Representations}\label{exp:analysis}

We next evaluate whether V-JEPA~2 provides greater downstream benefits than other visual representations commonly used in VLAs and broader embodied AI tasks.

\begin{table}[H]
  \caption{\textbf{LIBERO-Long task success rate (\%)} with different visual representations. Baseline denotes our implementation of the basic VLA.}
  \label{tab:dino_siglip_vjepa2}
  \begin{center}
    \begin{small}
      \begin{sc}
        \begin{tabular}{lcccccr}
          \toprule
          Model & Long SR (\%)\\
          \midrule
          Baseline & 25.8  \\
          Baseline+DINOv2    & 26.0 \\
          Baseline+DINOv2    & 31.2 \\
          Baseline+SigLIP    & 33.6 \\
          \textbf{Baseline+V-JEPA~2}  & \textbf{41.0} \\
          \bottomrule
        \end{tabular}
      \end{sc}
    \end{small}
    \vspace{-20pt}
  \end{center}
\end{table}

\paragraph{VLAs} Motivated by the analysis in Section~\ref{sec:analysis_of_features}, which suggests that static image-based representations may fail to capture key task-relevant states and effective policy priors, we conduct controlled replacement experiments by directly swapping the visual representation in the basic VLA baseline (Section~\ref{exp:implementation}). As shown in Table~\ref{tab:dino_siglip_vjepa2}, adapting DINOv2 yields marginal or even negative gains, while SigLIP leads to only modest improvements. In contrast, incorporating V-JEPA~2 results in substantially larger performance gains.

\vspace{-10pt}
\paragraph{Broader Embodied AI Tasks} Beyond the VLA backbones commonly adopted in robotic manipulation, we further compare V-JEPA~2 with strong pretrained visual representations (PVRs) evaluated in the broader embodied AI literature. Specifically, we consider CortexBench~\cite{majumdar2023we}, which comprises 17 tasks designed to assess the utility of pretrained visual representations for embodied control. We focus on 10 representative tasks and compare V-JEPA~2 against VC-1, a high-performing PVR reported in the same benchmark to outperform prior methods on average. Results in Tables~\ref{tab:cortexbench_1} and~\ref{tab:cortexbench_1} show remarkable improvements across the evaluated tasks: V-JEPA~2 outperforms VC-1 on most subtasks, indicating that video-predictive pretraining yields representations that transfer effectively to embodied decision making. Additional details are provided in Appendix~\ref{appendix:cortexbench}.

\section{Related Work}

\paragraph{Vision-Language-Action Models}
Building on the success of VLMs, recent VLAs have been able to process task instructions and visual observations and generate actions for manipulation. VLAs have made remarkable progress in language-conditioned control and generalization over task-specific controllers, enabling semantic reasoning and even achieving zero-shot performance on unseen tasks~\cite{zitkovich2023rt, yang2025instructvla, glossop2025cast, li2025switchvla}. However, despite advancements in instruction-following, current VLAs still struggle with insufficient visual understanding~\cite{sapkota2025vision, shao2025large}. Although several approaches have been proposed to improve visual understanding , including integrating explicit perception modules~\cite{yuan2025depthvla, huang2025tactile}, incorporating visual chain-of-thought reasoning~\cite{zhao2025cot, cen2025worldvla, cen2025rynnvla, zhong2025flowvla}, and utilizing methods based on 3D modeling and point clouds~\cite{lin2025evo, qu2025spatialvla}. However, these methods are often overly complex, difficult to implement in real-world settings or lack the desired effectiveness. Our goal is to identify a minimal and more general approach to enhance the visual understanding of current mainstream VLAs.

\vspace{-5pt}
\paragraph{Pretrained Vision Representations for Robotics.}
A growing body of work has developed pretrained visual representations tailored for robotics and manipulation \cite{nair2023r3m, mavip, xiao2022masked}. While these representations improve transfer in specific settings, they are often not sufficiently general to serve as universal backbones for modern VLA models. 
In practice, most VLM vision backbones used in recent VLAs (e.g., OpenVLA~\cite{kim2025openvla} and RDT~\cite{liu2024rdt}) are pretrained via either image-based self-supervised learning (e.g., the DINO family~\cite{caron2021emerging, oquab2024dinov2, simeoni2025dinov3}) or language--image contrastive learning (e.g., the SigLIP family~\cite{zhai2023sigmoid, tschannen2025siglip}).
Despite their success, these pretraining paradigms can be suboptimal for robotic environment understanding and lack of policy priors as discussed in Section~\ref{sec:intro}.
Motivated by these limitations, we explore pretrained video predictive representations for downstream robotic control.

\vspace{-5pt}
\paragraph{The Evolution of JEPA}  
The Joint-Embedding Predictive Architecture (JEPA) framework was first introduced to build predictive models of the world through self-supervised learning, enabling efficient understanding, prediction, and reasoning about the real world without relying on large labeled datasets or traditional generative objectives~\cite{lecun2022path}. This was followed by the development of I-JEPA~\cite{assran2023self}, V-JEPA~\cite{bardesrevisiting}, and V-JEPA~2~\cite{assran2025v}, demonstrating that this architecture is scalable and can generate visual representations that perform well across a wide range of tasks (e.g., motion understanding). Subsequent work has further shown that intuitive physics understanding emerges during the training of V-JEPA~\cite{garrido2025intuitive}, suggesting that the latent predictive learning approach inherent in V-JEPA produces broadly generalizable representations capable of understanding the physical world, predicting future states, and planning effectively in new situations. We aim to leverage these capabilities in robotics.

\section{Conclusion and Future Work}

We identify two critical forms of visual knowledge required by VLAs—environment understanding and policy priors—and show that widely used vision encoders fail to provide them, resulting in poor sample efficiency and weak generalization of VLAs. Crucially, we demonstrate that video-predictive embeddings, exemplified by V-JEPA~2, fill this gap by capturing task-relevant state representations and encoding temporal regularities that act as effective policy priors. Building on this finding, we introduce JEPA-VLA, a simple and general integration framework that consistently boosts VLA performance across multiple benchmarks.

This work opens several promising directions for future research. 
While we focus on a straightforward fusion strategy, more principled mechanisms for integrating predictive embeddings remain largely unexplored. 
More broadly, our findings suggest that leveraging large-scale video pretraining is a key ingredient for advancing visual understanding in robotics, and we hope this work will encourage further investigation into vision-centric approaches for building more robust and generalizable embodied agents.

\bibliography{example_paper}
\bibliographystyle{icml2026}

\newpage
\appendix
\onecolumn
\section{Details of Vision Representation Analysis}
\label{appendix:analysis}

This section provides implementation details for the analysis tasks introduced in Section~\ref{sec:analysis_of_features}. The hyperparameters used across all analysis experiments are summarized in Table~\ref{tab:hyper_analysis}. 

For all experiments, we split the dataset into training, validation, and test sets with a ratio of 8:1:1. Models are trained with early stopping: training terminates if the validation loss does not decrease for 10 consecutive epochs. The checkpoint with the lowest validation loss is selected for final evaluation on the test set.

\begin{table*}[h]
\caption{Hyperparameters used for vision representation analysis.}
\label{tab:hyper_analysis}
\begin{center}
\begin{small}
\begin{sc}
\begin{tabular}{lc}
\toprule
Hyperparameter & Value \\
\midrule
Attention Heads & 16 \\
Batch Size & 32 \\
Decoder Layers & 12 \\
Dropout Rate & 0.1 \\
Feedforward Dimension & 3072 \\
Learning Rate & $1 \times 10^{-4}$ \\
Maximum Epochs & 100 \\
Optimizer & AdamW \\
Weight Decay & $1 \times 10^{-5}$ \\
\bottomrule
\end{tabular}
\end{sc}
\end{small}
\end{center}
\vskip -0.1in
\end{table*}

\paragraph{Task-Relevant State Regression}
To evaluate whether visual representations encode task-relevant environment information, we regress the current relevant states from frozen visual features. Specifically, we prepend an additional \texttt{[CLS]} token to the representation sequence and use its output embedding for state regression. All relevant-state labels are normalized to have zero mean and unit variance to stabilize training.

\paragraph{Task-Irrelevant State Regression}
We assess sensitivity to task-irrelevant factors by regressing lighting parameters and background appearance.

For lighting regression, all lighting-related parameters (e.g., light direction, position, and diffuse components) are concatenated into a single target vector. An additional \texttt{[CLS]} token is introduced, and its embedding is used to predict the lighting parameters.

For background regression, we employ an additional linear upsampling head to reconstruct the background texture from the frozen visual representations. Reconstruction is supervised using the standard mean absolute error (MAE) loss.

\paragraph{Task-Relevant State Prediction}
To evaluate whether representations encode transition-aware information, we predict future changes in task-relevant states. Since control in LIBERO-10 is precise and state changes between adjacent time steps are often subtle, we predict the residual between the current state and the state after 10 time steps.

Unlike the regression setting, we do not normalize state values in this task, as normalization would distort the semantic meaning of residuals. Except for this difference, all other settings remain identical to those used in \emph{Task-Relevant State Regression}.

\begin{figure*}[ht]
  \vskip 0.2in
  \begin{center}
    \centerline{\includegraphics[width=\textwidth]{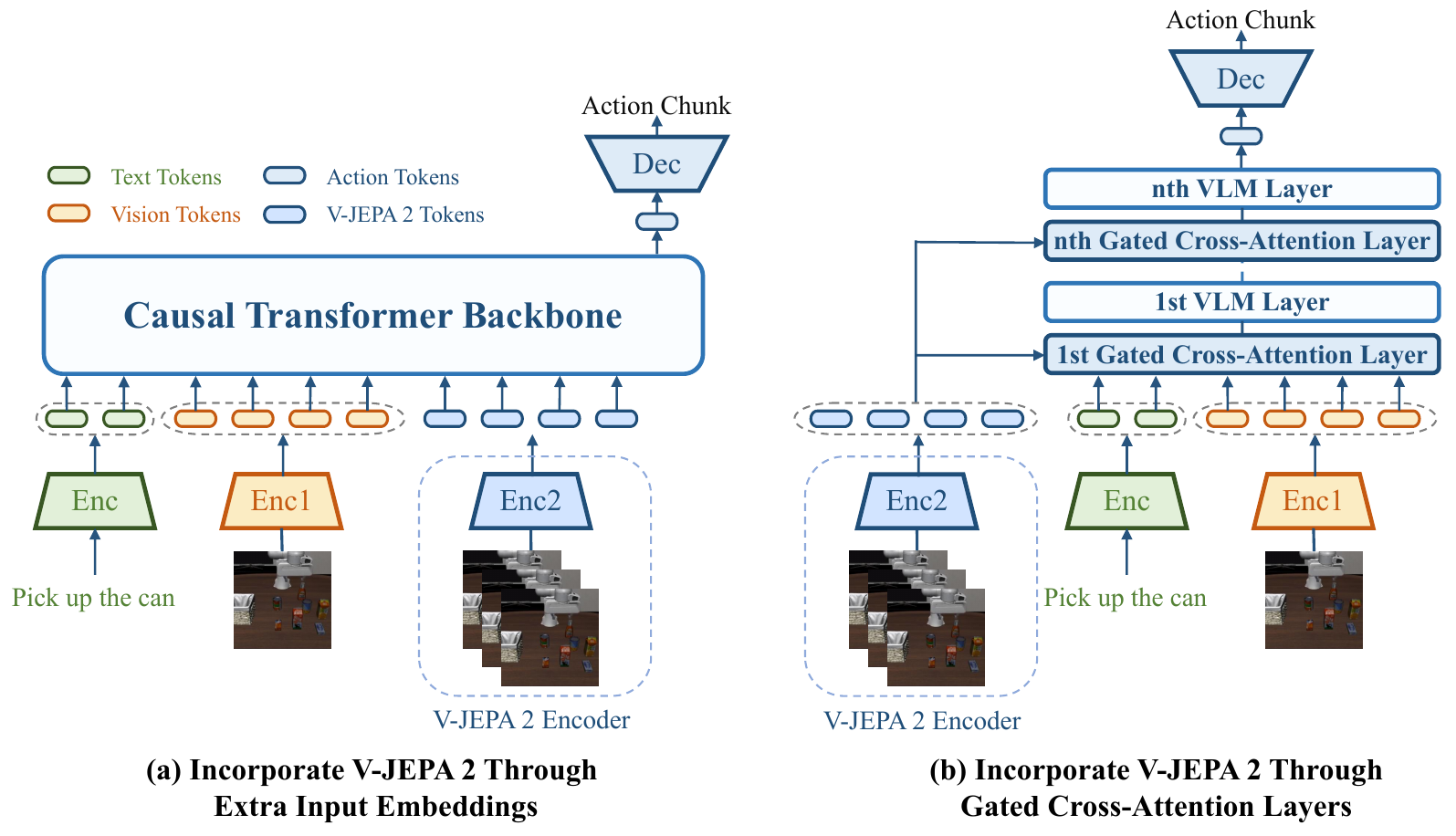}}
    \caption{\textbf{JEPA-VLA fusion architecture.} (a) For VLAs without large-scale robotic pretraining, V-JEPA~2 representations are concatenated as additional input embeddings. (b) For VLAs pretrained on large-scale robotic datasets, we integrate V-JEPA~2 representations using gated cross-attention, which enables adaptive fusion while preserving pretrained priors.}

    \label{fig:jepa_vla}
  \end{center}
\end{figure*}

\section{JEPA-VLA Fusion Method Details}\label{appendix:jepa_vla}

The overall JEPA-VLA fusion architecture is illustrated in Figure~\ref{fig:jepa_vla}. 
We consider two representative scenarios depending on whether the underlying Vision-Language-Action (VLA) model has been pretrained on large-scale robotic manipulation datasets.

For VLAs trained from scratch or without extensive robot-specific pretraining, a simple fusion strategy is sufficient. In this setting, V-JEPA~2 representations are treated as additional input embeddings and concatenated with the original token sequence. This straightforward design effectively injects video-predictive knowledge into the VLA, enabling the policy to leverage temporal and state-centric information provided by V-JEPA~2.

However, for VLAs that have already been pretrained on large-scale robotic datasets, naive concatenation can be detrimental. Directly introducing extra embeddings shifts the input distribution and may interfere with the pretrained representations, thereby disrupting the learned priors and leading to degraded performance. This issue is particularly pronounced when the pretrained VLA has already internalized strong task-specific or action-aligned representations.

To address this challenge, we incorporate V-JEPA~2 representations through gated cross-attention layers. In this design, the original VLA token embeddings serve as queries, while the V-JEPA~2 representations act as keys and values. The gating mechanism allows the model to adaptively control the contribution of predictive visual representations, selectively attending to V-JEPA~2 features when beneficial while preserving the original pretrained priors. This adaptive fusion enables effective knowledge transfer from V-JEPA~2 without destabilizing pretrained VLA representations.

\section{LIBERO-plus Benchmark Implementation Details}
\label{appendix:libero_plus}

Following the official LIBERO-plus evaluation protocol, each Vision-Language-Action (VLA) model is trained separately on the four LIBERO task suites. For each task suite, multiple checkpoints are saved during training, and the final performance is obtained by evaluating four selected checkpoints and reporting their average score. This protocol is designed to mitigate variance introduced by checkpoint selection.

In our experiments, we evaluate the same set of checkpoints reported in Table~\ref{tab:libero1-10}. We further report detailed LIBERO-plus results for each task suite separately, including Spatial, Object, Goal, and Long tasks. The corresponding results are presented in Tables~\ref{tab:libero_plus_spatial}, \ref{tab:libero_plus_object}, \ref{tab:libero_plus_goal}, and \ref{tab:libero_plus_long}, respectively.

\section{CortexBench Benchmark Implementation Details}
\label{appendix:cortexbench}

CortexBench is a comprehensive benchmark consisting of 17 tasks that span a wide range of robotic manipulation scenarios. While the benchmark provides reference results for several pretrained visual representations, we found that the reported performance of VC-1 could not be reliably reproduced for all tasks using the publicly released code and configurations.

As a result, we restrict our evaluation to a subset of 10 CortexBench tasks for which we are able to consistently reproduce baseline VC-1 results. On this subset, we compare the performance of V-JEPA~2 representations against VC-1 under identical experimental settings.

Furthermore, we note that the official CortexBench implementation reports task rewards rather than binary success rates for MetaWorld tasks. Since no explicit mapping from reward values to success criteria is provided, we report the reproduced reward scores for both VC-1 and V-JEPA~2 to ensure a fair and transparent comparison.

\section{Real-World Experiment Details}
\label{appendix:real_world}

\paragraph{Data Collection}
We manually collect a total of 100 demonstration trajectories for a real-world pick-and-place task using a single robotic arm. These trajectories are used to train both the baseline model and JEPA-VLA. To evaluate the effectiveness of JEPA-VLA in data-limited settings, we additionally train our method using only one-fifth of the collected data.

\paragraph{Training Details}
All models are trained for 30 epochs with an action chunk size of 10, a learning rate of $5 \times 10^{-6}$, and a batch size of 32. For JEPA-VLA, we extract representations from the V-JEPA~2 encoder using the two most recent visual frames. The resulting embeddings are concatenated with the original token sequence and provided as additional inputs to the VLA model.

\paragraph{Evaluation Details}
During evaluation, we adopt the attention masking mechanism proposed in WorldVLA~\cite{cen2025worldvla}, which prevents action tokens from attending to future actions. This masking strategy naturally supports parallel decoding during inference. We implement parallel decoding to accelerate evaluation and interpolate between consecutive actions to ensure smooth and coherent trajectories.

\begin{figure*}[ht]
  \vskip 0.2in
  \begin{center}
    \centerline{\includegraphics[width=\textwidth]{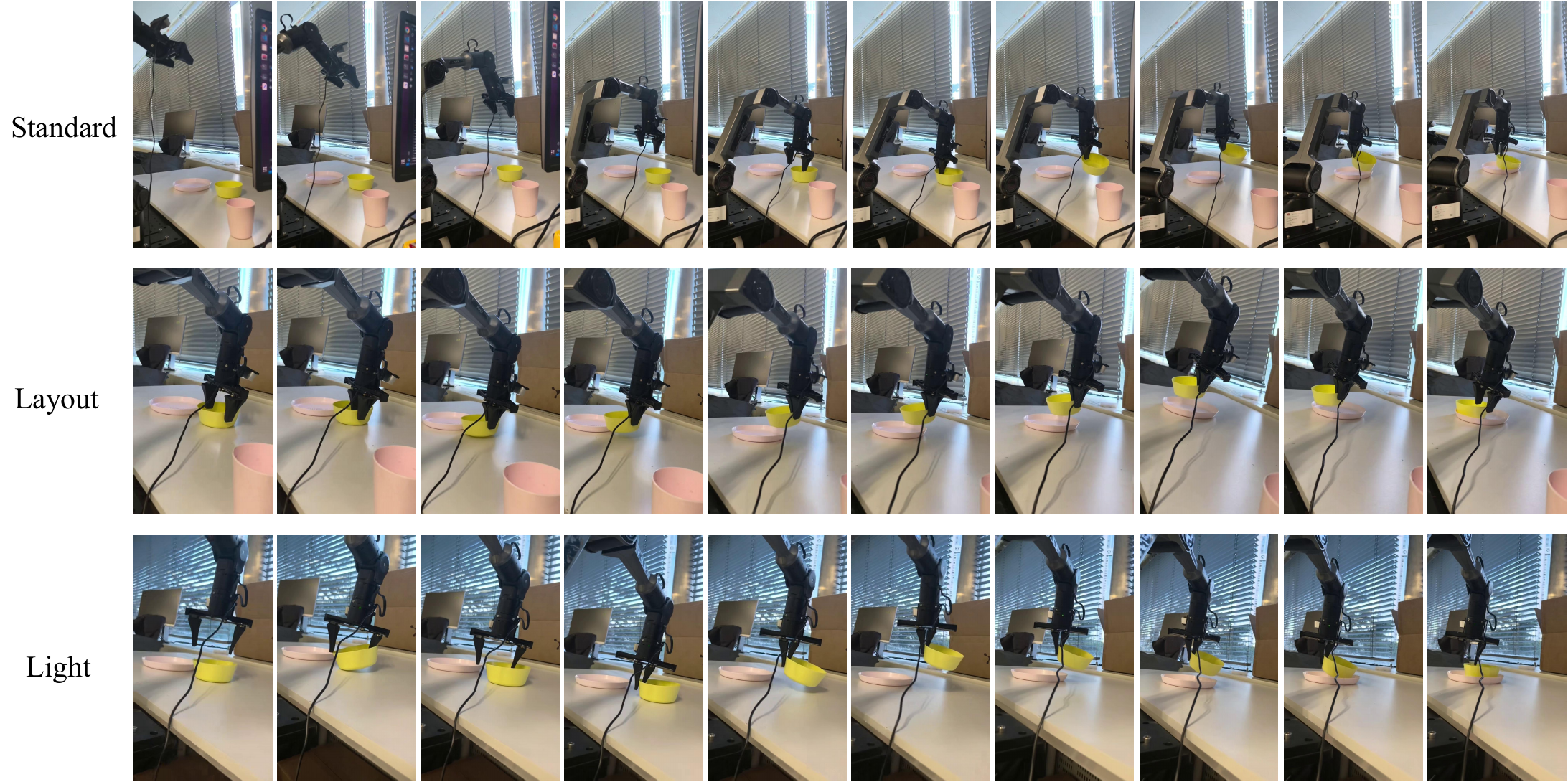}}
    \caption{Examples of real-world testing, respectively in standard, modified layouts and lights settings.}
    \label{fig:real_world_trj}
  \end{center}
\end{figure*}

Each model is evaluated over 10 independent trials. In addition to in-domain evaluation, we assess generalization performance under variations in lighting conditions and object layouts. Representative test scenarios and examples of successful trajectories are illustrated in Figure~\ref{fig:real_world_trj}.

\begin{table*}[t]
  \caption{LIBERO-plus task success rate(\%) of 1/10 LIBERO-Spatial training data.}
  \label{tab:libero_plus_spatial}
  \begin{center}
    \begin{small}
      \begin{sc}
        \begin{tabular}{lccccccccr}
          \toprule
          Model & Camera & Robot & Language & Light & Background & Noise & Layout & Total \\
          \midrule
          WorldVLA & 0.0 & 44.3 & 46.3 & 65.1 & 19.8 & 11.7 & 46.1 & 32.5\\
          Baseline & 0.0 & 24.9 & 51.5 & 44.2 & 10.5 & 1.7 & 31.4 & 23.8 \\
          \textbf{w/ours}  & \textbf{0.8} & \textbf{29.7} & \textbf{53.3} & \textbf{57.2} & \textbf{44.2} & \textbf{1.4} & \textbf{44.2} & \textbf{32.1}\\
          $\Delta$  & \textcolor[RGB]{17,159,73}{0.8} & \textcolor[RGB]{17,159,73}{4.8} & \textcolor[RGB]{17,159,73}{1.8} & \textcolor[RGB]{17,159,73}{13.0} & \textcolor[RGB]{17,159,73}{33.7} & \textcolor[RGB]{17,0,0}{-0.3} & \textcolor[RGB]{17,159,73}{12.8} & \textcolor[RGB]{17,159,73}{8.3}\\
          \bottomrule
        \end{tabular}
      \end{sc}
    \end{small}
  \end{center}
  \vskip -0.1in
\end{table*}

\begin{table*}[t]
  \caption{LIBERO-plus task success rate(\%) of 1/10 LIBERO-Object training data.}
  \label{tab:libero_plus_object}
  \begin{center}
    \begin{small}
      \begin{sc}
        \begin{tabular}{lccccccccr}
          \toprule
          Model & Camera & Robot & Language & Light & Background & Noise & Layout & Total \\
          \midrule
          WorldVLA & 0.0 & 26.4 & 57.2 & 20.5 & 17.3 & 18.0 & 53.6 & 28.6\\
          Baseline & 0.0 & 25.4 & 72.0 & 19.5 & 3.2 & 5.2 & 25.3 & 21.7 \\
          \textbf{w/ours}  & \textbf{0.0} & \textbf{33.0} & \textbf{91.5} & \textbf{35.0} & \textbf{7.3} & \textbf{6.4} & \textbf{34.7} & \textbf{29.6}\\
          $\Delta$  & \textcolor[RGB]{17,159,73}{0.0} & \textcolor[RGB]{17,159,73}{7.6} & \textcolor[RGB]{17,159,73}{19.5} & \textcolor[RGB]{17,159,73}{15.5} & \textcolor[RGB]{17,159,73}{4.1} & \textcolor[RGB]{17,159,73}{1.2} & \textcolor[RGB]{17,159,73}{9.4} & \textcolor[RGB]{17,159,73}{7.9}\\
          \bottomrule
        \end{tabular}
      \end{sc}
    \end{small}
  \end{center}
  \vskip -0.1in
\end{table*}

\begin{table*}[t]
  \caption{LIBERO-plus task success rate(\%) of 1/10 LIBERO-Goal training data.}
  \label{tab:libero_plus_goal}
  \begin{center}
    \begin{small}
      \begin{sc}
        \begin{tabular}{lccccccccr}
          \toprule
          Model & Camera & Robot & Language & Light & Background & Noise & Layout & Total \\
          \midrule
          WorldVLA & 0.3 & 30.6 & 42.2 & 68.8 & 30.3 & 13.5 & 47.4 & 31.8\\
          Baseline & 0.5 & 25.4 & 29.8 & 35.1 & 19.6 & 9.0 & 30.8 & 21.1 \\
          \textbf{w/ours}  & \textbf{0.5} & \textbf{28.9} & \textbf{39.5} & \textbf{40.5} & \textbf{36.3} & \textbf{7.7} & \textbf{36.0} & \textbf{26.2}\\
          $\Delta$  & \textcolor[RGB]{17,159,73}{0.0} & \textcolor[RGB]{17,159,73}{3.5} & \textcolor[RGB]{17,159,73}{9.7} & \textcolor[RGB]{17,159,73}{5.4} & \textcolor[RGB]{17,159,73}{16.7} & \textcolor[RGB]{17,0,0}{-1.3} & \textcolor[RGB]{17,159,73}{5.2} & \textcolor[RGB]{17,159,73}{5.1}\\
          \bottomrule
        \end{tabular}
      \end{sc}
    \end{small}
  \end{center}
  \vskip -0.1in
\end{table*}

\begin{table*}[t]
  \caption{LIBERO-plus task success rate(\%) of 1/10 LIBERO-Long training data.}
  \label{tab:libero_plus_long}
  \begin{center}
    \begin{small}
      \begin{sc}
        \begin{tabular}{lccccccccr}
          \toprule
          Model & Camera & Robot & Language & Light & Background & Noise & Layout & Total \\
          \midrule
          WorldVLA & 0.0 & 12.2 & 20.6 & 20.4 & 1.7 & 1.6 & 4.4 & 8.2\\
          Baseline & 0.0 & 7.4 & 22.5 & 6.2 & 7.3 & 0.7 & 19.9 & 8.7 \\
          \textbf{w/ours}  & \textbf{0.5} & \textbf{11.5} & \textbf{33.2} & \textbf{19.3} & \textbf{8.0} & \textbf{0.9} & \textbf{16.4} & \textbf{12.1}\\
          $\Delta$  & \textcolor[RGB]{17,159,73}{0.5} & \textcolor[RGB]{17,159,73}{4.1} & \textcolor[RGB]{17,159,73}{10.7} & \textcolor[RGB]{17,159,73}{13.1} & \textcolor[RGB]{17,159,73}{0.7} & \textcolor[RGB]{17,159,73}{0.2} & \textcolor[RGB]{17,0,0}{-3.5} & \textcolor[RGB]{17,159,73}{3.4}\\
          \bottomrule
        \end{tabular}
      \end{sc}
    \end{small}
  \end{center}
  \vskip -0.1in
\end{table*}

\end{document}